\def\year{2020}\relax
\documentclass[letterpaper]{article} 
\usepackage{aaai20}  
\usepackage{times}  
\usepackage{helvet} 
\usepackage{courier}  
\usepackage[hyphens]{url}  
\usepackage{graphicx} 
\def\UrlFont{\rm}  
\usepackage{graphicx}  
\usepackage{amsmath}
\usepackage{amsthm}
\usepackage{amssymb}
\usepackage{algorithm}
\usepackage{algorithmic}
\usepackage{subcaption}
\usepackage{booktabs}

\usepackage[makeroom]{cancel}
\DeclareRobustCommand
  \Fixedcdots{\mathinner{{\cdot}\mkern1mu{\cdot}\mkern1mu{\cdot}}}
  
\DeclareRobustCommand
  \Fixedldots{\mathinner{{.}\mkern1mu{.}\mkern1mu{.}}}

\usepackage{amsfonts}
\usepackage{bm}
\newcommand{\argmin}{\mathop{\mathrm{argmin}}}

\usepackage[sectionbib]{bibunits}
\usepackage{pdfpages}

\title{Learning to Communicate Implicitly by Actions}
\author{Zheng Tian$^\dagger$, Shihao Zou$^\dagger$, Ian Davies$^\dagger$, Tim Warr$^\dagger$, Lisheng Wu$^\dagger$, Haitham Bou Ammar$^\dagger$ $^\ddagger$, Jun Wang$^\dagger$\\
$^\dagger$ University College London \\
$^\ddagger$ Huawei R$\&$D UK\\
\{zheng.tian.11, shihao.zou.17, ian.davies.12, tim.warr.17, lisheng.wu.17\}@ucl.ac.uk\\
haitham.bouammar@huawei.com\\
jun.wang@cs.ucl.ac.uk
}
\begin{document}
\maketitle

\begin{abstract}
In situations where explicit communication is limited, human collaborators act by learning to: (i) infer meaning behind their partner's actions, and (ii) convey private information about the state to their partner implicitly through actions. The first component of this learning process has been well-studied in multi-agent systems, whereas the second --- which is equally crucial for successful collaboration --- has not. To mimic both components mentioned above, thereby completing the learning process, we introduce a novel algorithm: Policy Belief Learning (PBL). PBL uses a belief module to model the other agent's private information and a policy module to form a distribution over actions informed by the belief module. Furthermore, to encourage communication by actions, we propose a novel auxiliary reward which incentivizes one agent to help its partner to make correct inferences about its private information. The auxiliary reward for communication is integrated into the learning of the policy module. We evaluate our approach on a set of environments including a matrix game, particle environment and the non-competitive bidding problem from contract bridge. We show empirically that this auxiliary reward is effective and easy to generalize. These results demonstrate that our PBL algorithm can produce strong pairs of agents in collaborative games where explicit communication is disabled.
\end{abstract}

\section{Introduction} \label{intro}
In collaborative multi-agent systems, communication is essential for agents to learn to behave as a collective rather than a collection of individuals. This is particularly important in the imperfect-information setting, where private information becomes crucial to success.  In such cases, efficient communication protocols between agents are needed for private information exchange, coordinated joint-action exploration, and true world-state inference.

In typical multi-agent reinforcement learning (MARL) settings, designers incorporate explicit communication channels hoping to conceptually resemble language or verbal communication which are known to be important for human interaction~\cite{baker1999role}. Though they can be used for facilitating collaboration in MARL, explicit communication channels come at additional computational and memory costs, making them difficult to deploy in decentralized control~\cite{roth2006communicate}.

Environments where explicit communication is difficult or prohibited are common. These settings can be synthetic such as those in games, e.g., bridge and Hanabi, but also frequently appear in real-world tasks such as autonomous driving and autonomous fleet control. In these situations, humans rely upon implicit communication as a means of information exchange \cite{rasouli2017agreeing} and are effective in learning to infer the implicit meaning behind others' actions~\cite{heider1944experimental}.
The ability to perform such inference requires the attribution of a mental state and reasoning mechanism to others. This ability is known as theory of mind \cite{premack_woodruff_1978}. In this work, we develop agents that benefit from considering others' perspectives and thereby explore the further development of machine theory of mind~\cite{Rabinowitz2018}.

Previous works have considered ways in which an agent can, by observing an opponent's behavior, build a model of opponents' characteristics, objectives or hidden information  either implicitly~\cite{He2016,Bard2013} or explicitly \cite{Raileanu2018,li2018dynamic}. Whilst these works are of great value, they overlook the fact that an agent should also consider that it is being modeled and adapt its behavior accordingly, thereby demonstrating a theory of mind. For instance, in collaborative tasks, a decision-maker could choose to take actions which are informative to its teammates, whereas, in competitive situations, agents may act to conceal private information to prevent their opponents from modeling them effectively.

In this paper, we propose a generic framework, titled policy belief learning (PBL), for learning to cooperate in imperfect information multi-agent games. Our work combines opponent modeling with a policy that considers that it is being modeled. PBL consists of a \textit{belief module}, which models other agents' private information by considering their previous actions, and a \textit{policy module} which combines the agent's current observation with their beliefs to return a distribution over actions. We also propose a novel auxiliary reward for encouraging communication by actions, which is integrated into PBL. Our experiments show that agents trained using PBL can learn collaborative behaviors more effectively than a number of meaningful baselines without requiring any explicit communication. We conduct a complete ablation study to analyze the effectiveness of different components within PBL in our bridge experiment.

\section{Related Work}
\label{relw}

Our work is closely related to \cite{Albrecht2017,lowe2017multi,Mealing,Raileanu2018} where agents build models to estimate other agents' hidden information. Contrastingly, our work enhances a ``flat'' opponent model with recursive reasoning. ``Flat'' opponent models estimate only the hidden information of opponents. Recursive reasoning requires making decisions based on the mental states of others as well as the state of the environment. In contrast to works such as I-POMDP~\cite{Gmytrasiewicz:2005} and PR2~\cite{wen2018probabilistic} where the nested belief is embedded into the training agent's opponent model, we incorporate level-1 nested belief ``I believe that you believe" into our policy by a novel auxiliary reward.

Recently, there has been a surge of interest in using reinforcement learning (RL) approaches to learn communication protocols \cite{FoersterAFW16a,LazaridouPB16b,MordatchA17,SukhbaatarSF16}. Most of these works enable agents to communicate via an explicit channel. Among these works, \citeauthor{MordatchA17}~\shortcite{MordatchA17} also observe the emergence of non-verbal communication in collaborative environments without an explicit communication channel, where agents are exclusively either a sender or a receiver. Similar research is also conducted in \cite{DEWEERD201510}. In our setting, we do not restrict agents to be exclusively a sender or a receiver of communications -- agents can communicate mutually by actions. ~\citeauthor{Knepper:2017:ICJ:2909824.3020226}~\shortcite{Knepper:2017:ICJ:2909824.3020226} propose a framework for implicit communication in a cooperative setting and show that various problems can be mapped into this framework. Although our work is conceptually close to \cite{Knepper:2017:ICJ:2909824.3020226}, we go further and present a practical algorithm for training agents. The recent work of~\cite{FOERSTER2018BAD} solves an imperfect information problem as considered here from a different angle. We approach the problem by encouraging agents to exchange critical information through their actions whereas \citeauthor{FOERSTER2018BAD} train a public player to choose an optimal deterministic policy for players in a game based on publicly observable information. In Cooperative Inverse RL (CIRL) where robotic agents try to infer a human's private reward function from their actions \cite{hadfield2016cooperative}, optimal solutions need to produce behavior that coveys information.

\citeauthor{dragan2013legibility} \shortcite{dragan2013legibility} consider how to train agents to exhibit legible behavior (i.e. behavior from which it is easy to infer the intention). Their approach is dependent on a hand-crafted cost function to attain informative behavior. Mutual information has been used as a means to promote coordination without the need for a human engineered cost function. \citeauthor{Strouse:2018:LSH:3327546.3327688} \shortcite{Strouse:2018:LSH:3327546.3327688} use a mutual information objective to encourage an agent to reveal or hide its intention. In a related work, \citeauthor{jaques2019social} \shortcite{jaques2019social} utilize a mutual information objective to imbue agents with social influence. While the objective of maximal mutual information in actions can yield highly effective collaborating agents, a mutual information objective in itself is insufficient to necessitate the development of implicit communication by actions. \citeauthor{eccles2019learning} \shortcite{eccles2019learning} introduce a reciprocity reward as an alternative approach to solve social dilemmas..

A distinguishing feature of our work in relation to previous works in multi-agent communication is that we do not have a predefined explicit communication protocol or learn to communicate through an explicit channel. Information exchange can only happen via actions. In contrast to previous works focusing on unilaterally making actions informative, we focus on bilateral communication by actions where information transmission is directed to a specific party with potentially limited reasoning ability.
Our agents learn to communicate through iterated policy and belief updates such that the resulting communication mechanism and belief models are interdependent. The development of a communication mechanism therefore requires either direct access to the mental state of other agents (via centralized training) or the ability to mentalize, commonly known as theory of mind. We investigate our proposed algorithm in both settings.

\section{Problem Definition} \label{model}
We consider a set of agents, denoted by $\mathcal{N}$, interacting with an unknown environment by executing actions from a joint set $\mathcal{A}=\{\mathcal{A}_{1}, \Fixedldots, \mathcal{A}_{N}\}$, with $\mathcal{A}_{i}$ denoting the action space of agent $i$, and $N$ the total number of agents. To enable models that approximate real-world scenarios, we assume private and public information states. Private information states, jointly (across agents) denoted by $\mathcal{X}= \{\mathcal{X}_{1}, \Fixedldots, \mathcal{X}_{N}\}$ are a set of hidden information states where $\mathcal{X}_{i}$ is only observable by agent $i$, while public states $\mathcal{O}$ are observed by all agents. We assume that hidden information states at each time step are sampled from an unknown distribution $\mathcal{P}_{x}: \mathcal{X} \rightarrow [0,1]$, while public states evolve from an initial distribution $\mathcal{P}_{o}: \mathcal{O} \rightarrow [0, 1]$, according to a stochastic transition model $\mathcal{T}: \mathcal{O} \times \mathcal{X} \times \mathcal{A}_{1} \times \Fixedcdots \times \mathcal{A}_{N} \times \mathcal{O} \rightarrow [0,1]$. Having transitioned to a successor state according to $\mathcal{T}$, agents receive rewards from $\mathcal{R}: \mathcal{S}\times \mathcal{A}_{1}\times \Fixedcdots \times \mathcal{A}_{N}\rightarrow \mathbb{R}$, where we have used $\mathcal{S} = \mathcal{O} \times \mathcal{X}$ to denote joint state descriptions that incorporate both public and private information. Finally, rewards are discounted over time by a factor $\gamma \in (0, 1]$.
With this notation, our problem can be described succinctly by the tuple: $\left \langle \mathcal{N}, \mathcal{A}, \mathcal{O}, \mathcal{X}, \mathcal{T}, \mathcal{R}, \mathcal{P}_{x}, \mathcal{P}_{o}, \gamma \right\rangle$, which we refer to as an imperfect-information Markov decision process (I2MDP)\footnote{We also note that our problem can be formalized as a decentralized partially observable Markov decision process (Dec-POMDP)~\cite{Bernstein2013DECPOMDP}.}. In this work, we simplify the problem by assuming that hidden information states $\mathcal{X}$ are temporally static and are given at the beginning of the game.

We interpret the joint policy from the perspective of agent $i$ such that $\pi = (\pi^{i}(a^{i}| s), \pi^{-i}(a^{-i}| s))$, where $\pi^{-i}(a^{-i}| s)$ is a compact representation of the joint policy of all agents excluding agent $i$. In the collaborative setting, each agent is presumed to pursue the shared maximal cumulative reward expressed as
\begin{align}
\max~~\eta^i(\pi) = \mathbb { E } \left[ \sum _ { t = 1 } ^ { \infty } \gamma ^ { t } R( s_{t}, a^{i} _ { t }, a^{-i} _ { t  }  ) \right],
\label{marl_target}
\end{align}
where $s_t=[x^i_t, x^{-i}_t, o_t]$ is the current full information state, $(a^i_t, a^{-i}_t)$ are joint actions taken by agent $i$ and all other agents respectively at time $t$ and $\gamma$ is a discount factor. 

\section{Policy Belief Learning}
\begin{figure*}[ht]
\centering
\begin{subfigure}{0.35\textwidth}
\centering
\includegraphics[height=4.5cm]{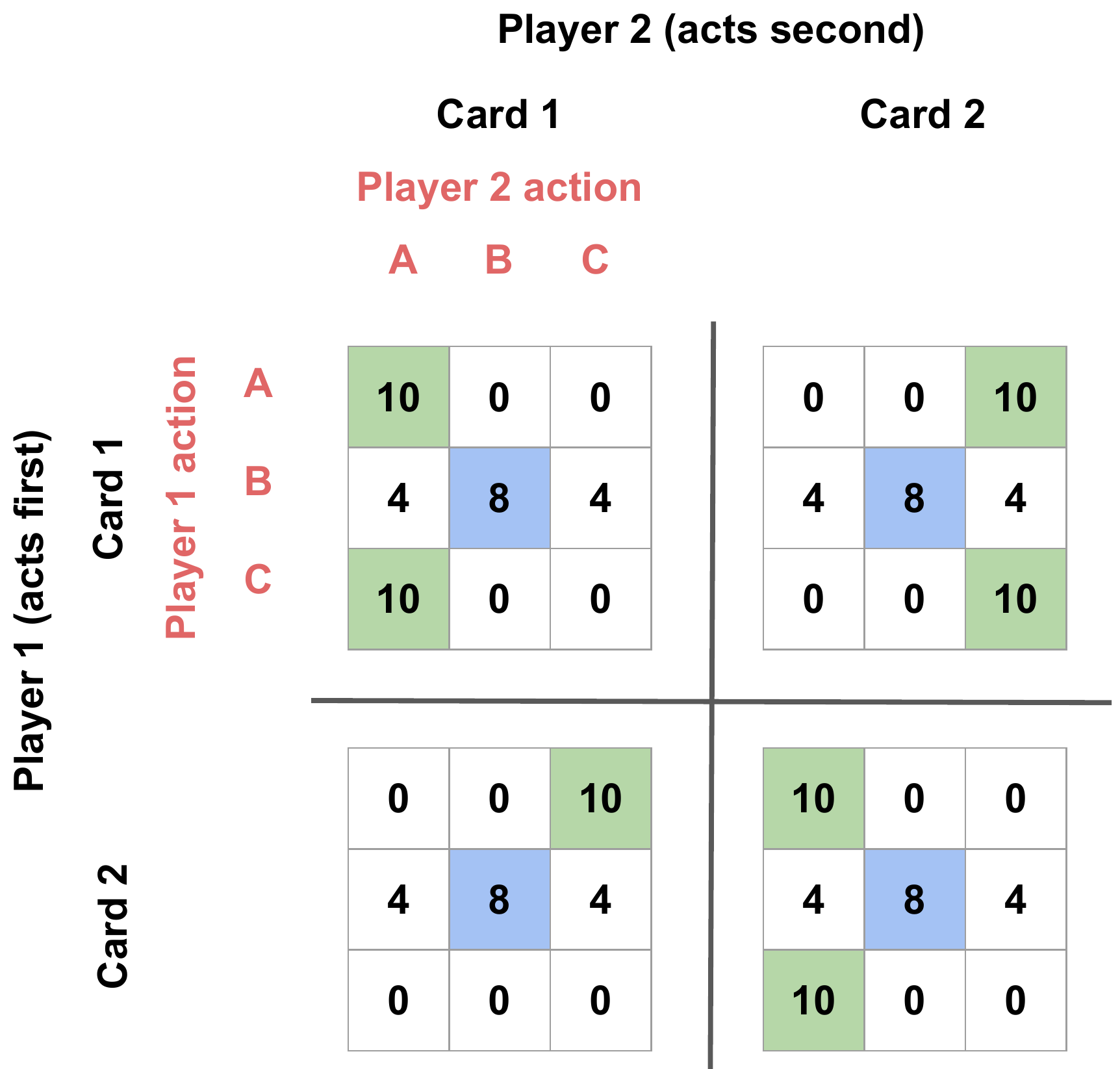}
\caption{Payoff for the matrix game}
\label{fig:matrixgametable}
\end{subfigure}
\hspace{1cm}
\begin{subfigure}{0.5\textwidth}
\centering
\includegraphics[height=4.5cm]{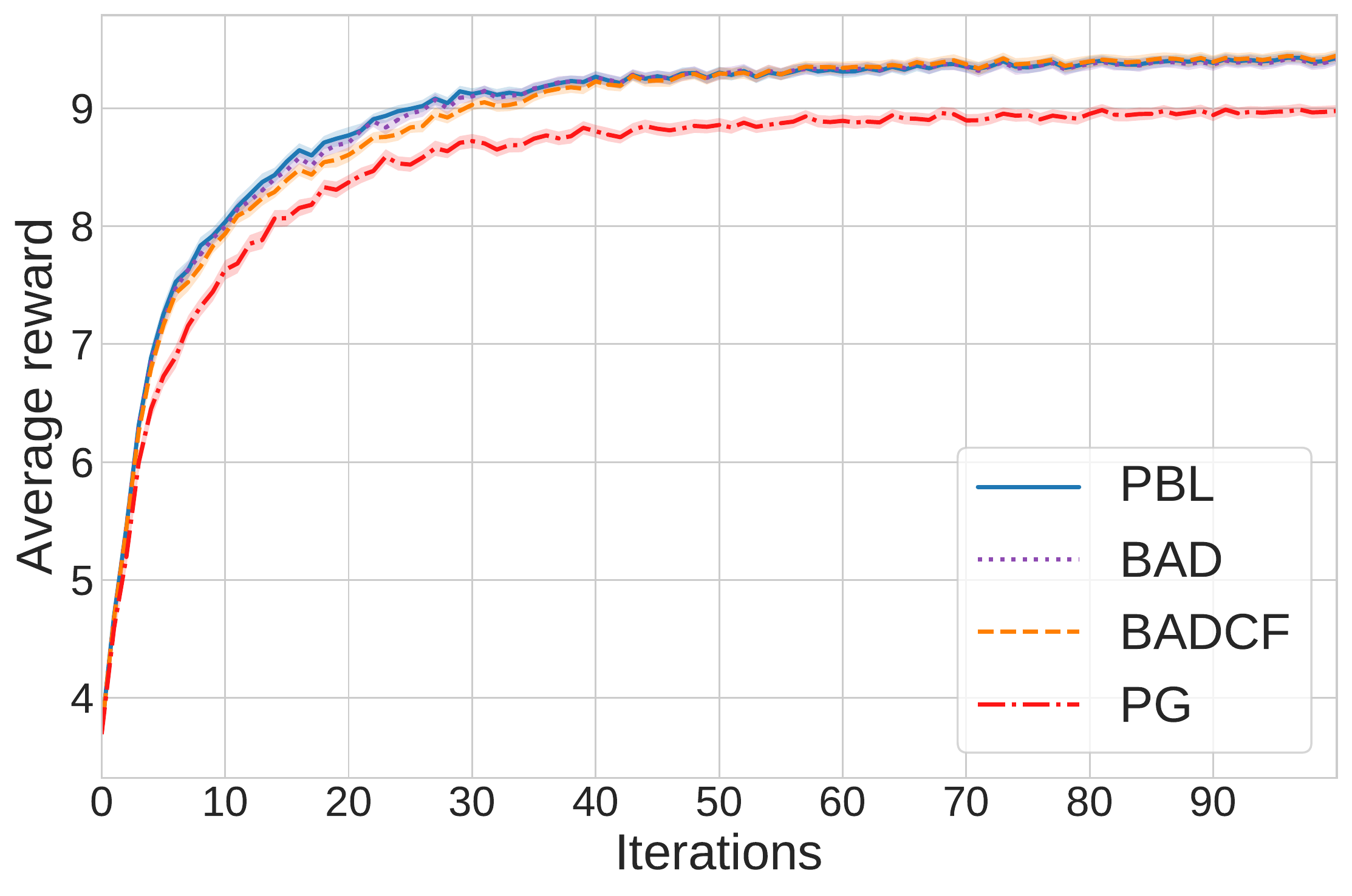}
\caption{Learning curves of PBL and baselines over 100 runs}
\label{fig:matrixgamecurve}
\end{subfigure}
\caption{Matrix game experiment and results.}
\label{fig:matrixgame}
\end{figure*}

Applying naive single agent reinforcement learning (SARL) algorithms to our problem will lead to poor performance. One reason for this is the partial observability of the environment. To succeed in a partially observable environment, an agent is often required to maintain a belief state. Recall that, in our setting, the environment state is formed from the union of the private information of all agents and the publicly observable information,  $s_t=[x^i_t, x^{-i}_t, o_t]$. We therefore learn a belief module $\Phi^i(x^{-i}_t)$ to model other agents' private information $x^{-i}_t$ which is the only hidden information from the perspective of agent $i$ in our setting. We assume that an agent can model $x_{t}^{-i}$ given the history of public information and actions executed by other agents $h_{t}^{i} = \{o_{1:t-1}, a_{1:t-1}^{-i}\}$. We use a NN to parameterize the belief module which takes in the history of public information and produces a belief state $b^i_t=\Phi^i(x^{-i}_t|h_{t}^{i})$. The belief state together with information observable by agent $i$ forms a sufficient statistic, $\hat{s}^i_t=[x^i_t, b^i_t, o_t]$, which contains all the information necessary for the agent to act optimally~\cite{Astrom:1965}. We use a separate NN to parameterize agent $i$'s policy $\pi^i(a^i_t|\hat{s}^i_t)$ which takes in the estimated environment state $\hat{s}^i_t$ and outputs a distribution over actions. As we assume hidden information is temporally static, we will drop the time script for it in the rest of the paper.

The presence of multiple learning agents interacting with the environment renders the environment non-stationary. This further limits the success of SARL algorithms which are generally designed for environments with stationary dynamics. To solve this, we adopt centralized training and decentralized execution, where during training all agents are recognized as one central representative agent differing only by their observations. Under this approach, one can imagine belief models  $\Phi^i(x^{-i}|h^{i}_{t})$ and $\Phi^{-i}(x^i|h^{-i}_t)$ sharing parameters $\bm{\phi}$. The input data, however, varies across agents due to the dependency on both $h_{t}^{i}$ and $h_{t}^{-i}$. 
In a similar fashion, we let policies share the parameters $\bm{\theta}$. Consequently, one may think of updating $\bm{\theta}$ and $\bm{\phi}$ using one joint data set aggregated across agents. Without loss of generality, in the remainder of this section, we discuss the learning procedure from the point of view of a single agent, agent $i$. 

We first present the learning procedure of our belief module. At iteration $k$, we use the current policy $\pi_{[k]}(a^i|\hat{s})$ to generate a data set of size $M$, $\Omega_{[k]}=\{(x^{-i}_{j}, h^i_{j})_{j=1}^{M}\}$, using self-play and learn a new belief module by minimizing: 
\begin{align}
\label{Eq:Phi}
\bm{\phi}_{[k]} := \argmin_{\bm{\phi}} \mathbb{E}_{ (x^{-i},h^i){\sim}\Omega_{[k-1]}}\left[\text{KL}(x_{
j}^{-i}||b^i_j(h_j^i;\bm{\phi})\right],
\end{align}
where $\text{KL}(\cdot||\cdot)$ is the Kullback--Leibler(KL) divergence and we use a one-hot vector to encode the ground truth, $x_{j}^{-i}$, when we calculate the relevant KL-divergence. 

With updated belief module $\Phi_{[k]}$, we learn a new policy for the next iteration, $\pi_{[k+1]}$, via a policy gradient algorithm. Sharing information in multi-agent cooperative games through communication reduces intractability by enabling coordinated behavior. Rather than implementing expensive protocols~\cite{heider1944experimental}, we encourage agents to \emph{implicitly communicate} through actions by introducing a novel auxiliary reward signal. To do so, notice that in the centralized setting agent $i$ has the ability to consult its opponent's belief model $\Phi^{-i}(x^{i}|h^{-i}_{t})$ thereby exploiting the fact that other agents hold beliefs over its private information $x_{i}$.
In fact, comparing $b^{-i}_t$ to the ground-truth $x^{i}$ enables agent $i$ to learn which actions bring these two quantities closer together and thereby learn informative behavior. This can be achieved through an auxiliary reward signal devised to encourage informative action communication:
\begin{equation}
\label{eq:centralisedcommunication}
    r^i_{c,t} =  \text{KL}(x^{i}||b^{-i,*})
    - \text{KL}(x^{i}||b^{-i}_{t+1}),
\end{equation}
where $b^{-i, *}=\Phi^{-i}_{[k]}(x^{i}|h_{t,*}^{-i})$ is agent $-i$'s
best belief 
(so-far) about agent $i$'s private information:
\begin{equation*}
       b^{-i,*}=\argmin~ \text{KL}(x^{i}||b_u^{-i})\ \forall\ u\leq t .
\end{equation*}
In other words, $r^i_{c,t}$ encourages communication as it is proportional to the improvement in the opponent's belief (for a fixed belief model $\Phi^{-i}_{[k]}(x^{i}|h_{t+1}^{-i})$), measured by its proximity to the ground-truth, resulting from the opponent observing agent $i$'s action $a^{i}_{t}$. Hence, during the policy learning step of PBL, we apply a policy gradient algorithm with a shaped reward of the form:\footnote{Please note, we omit the agent index $i$ in the reward equation, as we shape rewards similarly for all agents.}
\begin{align}
\label{Eq:RewardsHere}
    r = r_e + \alpha r_c ,
\end{align}
where $r_e$ is the reward from the environment, $r_c$ is the communication reward and $\alpha \geq 0$ balances the communication and environment rewards. 

\begin{algorithm}[ht]
\caption{Per-Agent Policy Belief Learning (PBL)}
\label{Algo:PBL}
\begin{algorithmic}[1]
        \STATE \textbf{Initialize:} Randomly initialize policy $\pi_{0}$ and belief $\Phi_{0}$
        \STATE Pre-train $\pi_{0}$
        \FOR{$k=0$ to $\text{max\_iterations}$}
        \STATE Sample episodes for belief training using self-play forming the data set $\Omega_{[k]}$
        \STATE Update belief network using data from $\Omega_{[k]}$ solving Equation~\ref{Eq:Phi}
        \STATE Given updated beliefs $\Phi_{[k+1]}(\cdot)$, update policy $\pi(\cdot)$ (policy gradients with rewards from Equation~\ref{Eq:RewardsHere})
        \ENDFOR
        \STATE \textbf{Output:} Final policy, and belief model
    \end{algorithmic}
\end{algorithm}

Initially, in the absence of a belief module, we pre-train a policy $\pi_{[0]}$ naively by ignoring the existence of other agents in the environment. As an agent's reasoning ability may be limited, we may then iterate between Belief and Policy learning multiple times until either the allocated computational resources are exhausted or the policy and belief modules converge. We summarize the main steps of PBL in Algorithm~\ref{Algo:PBL}. Note that, although information can be leaked during training, as training is centralized, distributed test-phase execution ensures hidden-private variables during execution. 

\section{Machine Theory of Mind}
In PBL, we adopt a centralized training and decentralized execution scheme where agents share the same belief and policy models. In reality, however, it is unlikely that two people will have exactly the same reasoning process. In contrast to requiring everyone to have the same reasoning process, a person's success in navigating social dynamics relies on their ability to attribute mental states to others. This attribution of mental states to others is known as theory of mind~\cite{premack_woodruff_1978}. Theory of mind is fundamental to human social interaction which requires the recognition of other sensory perspectives, the understanding of other mental states, and the recognition of complex non-verbal signals of emotional state~\cite{Lemaignan2015MutualMI}. In collaboration problems without an explicit communication channel, humans can effectively establish an understanding of each other's mental state and subsequently select appropriate actions. For example, a teacher will reiterate a difficult concept to students if she infers from the students' facial expressions that they have not understood. The effort of one agent to model the mental state of another is characterized as Mutual Modeling~\cite{dillenbourg:hal-00190240}.

In our work, we also investigate whether the proposed communication reward can be generalized to a distributed setting which resembles a human application of theory of mind. Under this setting, we train a separate belief model for each agent so that $\Phi^i(x^{-i}|h^{i}_{t})$ and $\Phi^{-i}(x^i|h^{-i}_t)$ do not share parameters $(\phi^i\neq\phi^{-i})$. Without centralization, an agent can only measure how informative its action is to others with its own belief model. Assuming agents can perfectly recall their past actions and observations, agent $i$ computes its communication reward as:\footnote{Note the difference of super/sub-scripts of the belief model and its parameters when compared to Equation~\ref{eq:centralisedcommunication}.}
\begin{equation*}
    r^i_{c,t} =  \text{KL}(x^{i}||\tilde{b}_t^{i,*})
    - \text{KL}((x^{i}||\tilde{b}^i_{t+1}),
\end{equation*}
where $\tilde{b}^{i,*}_t = \Phi^i(x^i|h^{-i}_{t,*})$ and $\tilde{b}^{i}_{t+1} = \Phi^i(x^i|h^{-i}_{t+1})$.
In this way, an agent essentially establishes a mental state of others with its own belief model and acts upon it. We humbly believe this could be a step towards machine theory of mind where algorithmic agents learn to attribute mental states to others and adjust their behavior accordingly.

The ability to mentalize relieves the restriction of collaborators having the same reasoning process. However, the success of collaboration still relies on the correctness of one's belief about the mental states of others. For instance, correctly inferring other drivers' mental states and conventions can reduce the likelihood of traffic accidents. Therefore road safety education is important as it reduces variability among drivers reasoning processes. In our work, this alignment amounts to the similarity between two agents' trained belief models which is affected by training data, initialization of weights, training algorithms and so on. We leave investigation of the robustness of collaboration to variability in collaborators' belief models to future work.

\section{Experiments \& Results}
\label{exp}
We test our algorithms in three experiments. In the first, we validate the correctness of the PBL framework which integrates our communication reward with iterative belief and policy module training in a simple matrix game. In this relatively simple experiment, PBL achieves near optimal performance. Equipped with this knowledge, we further apply PBL to the non-competitive bridge bidding problem to verify its scalability to more complex problems. Lastly, we investigate the efficacy of the proposed communication reward in a distributed training setting. 
\begin{figure*}[t]
\centering
\begin{subfigure}{0.5\textwidth}
\centering
\includegraphics[height=4.5cm]{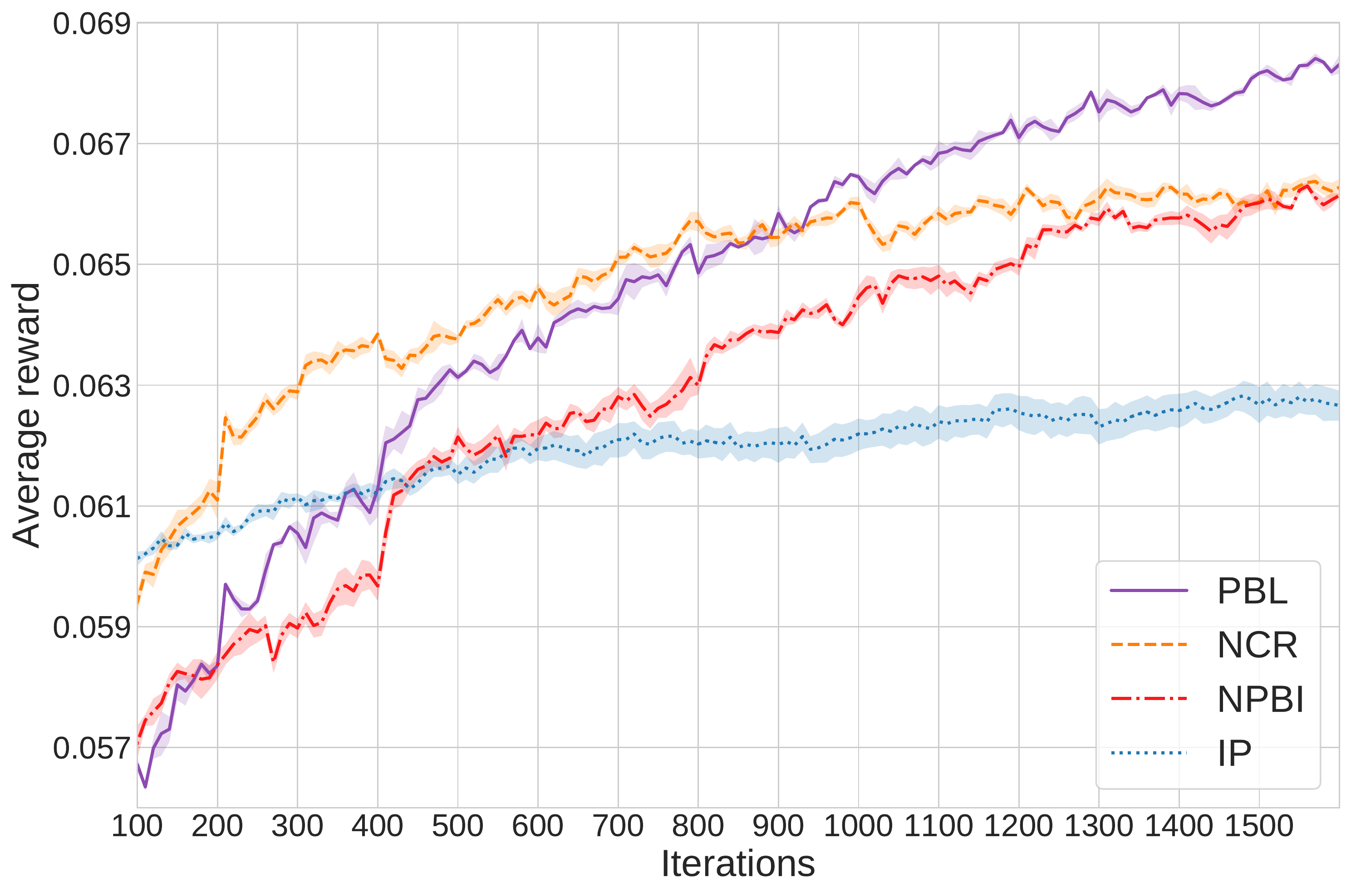}
\caption{Learning curves for non-competitive bridge bidding}
\label{fig:non-comp-bridge}
\end{subfigure}
~~
\begin{subfigure}{0.44\textwidth}
\centering
\includegraphics[height=4.5cm]{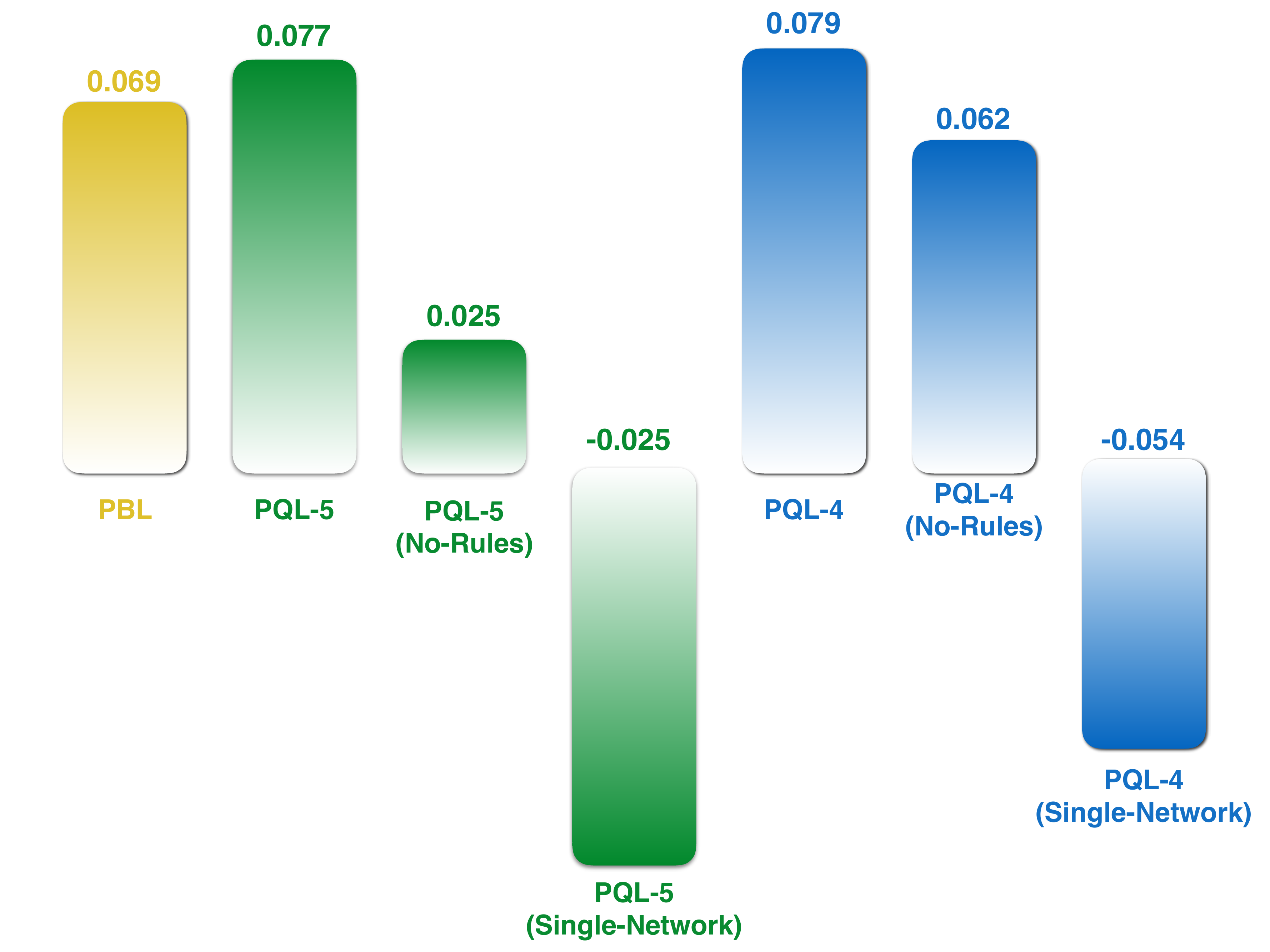}
\caption{Comparison of PBL and PQL}
\label{fig:pbl_vs_pql_bar}
\end{subfigure}
\caption{a) Learning curves for non-competitive bridge bidding with a warm start from a model trained to predict the score distribution (average reward at warm start: 0.038). Details of warm start provided in Appendix \ref{appendix:warmstart}. b) Bar graph comparing PBL to variants of PQL, with the full version of PQL results as reported in~\cite{Yeh16}.}
\label{fig:bridgegame}
\end{figure*}

\subsection{Matrix Game}
We test our PBL algorithm on a matrix card game where an implicit communication strategy is required to achieve the global optimum. This game is first proposed in \cite{FOERSTER2018BAD}. There are two players and each player receives a card drawn from $\{\text{card 1}, \text{card 2}\}$ independently at the beginning of the game. Player 1 acts first and Player 2 responds after observing Player 1's action. Neither player can see the other's hand. By the design of the payoff table (shown in Figure.~\ref{fig:matrixgametable}), Player 1 has to use actions C and A to signify that it holds Cards 1 and 2 respectively so that Player 2 can choose its actions optimally with the given information. We compare PBL with algorithms proposed in \cite{FOERSTER2018BAD} and vanilla policy gradient. As can be seen from Figure \ref{fig:matrixgamecurve}, PBL performs similarly to BAD and BAD-CF on this simple game and outperforms vanilla policy gradient significantly. This demonstrates a proof of principle for PBL in a multi-agent imperfect information coordination game.

\subsection{Contract Bridge Case-Study}\label{contractbridge}

\begin{figure*}[ht]
\centering
\includegraphics[width=0.87\textwidth]{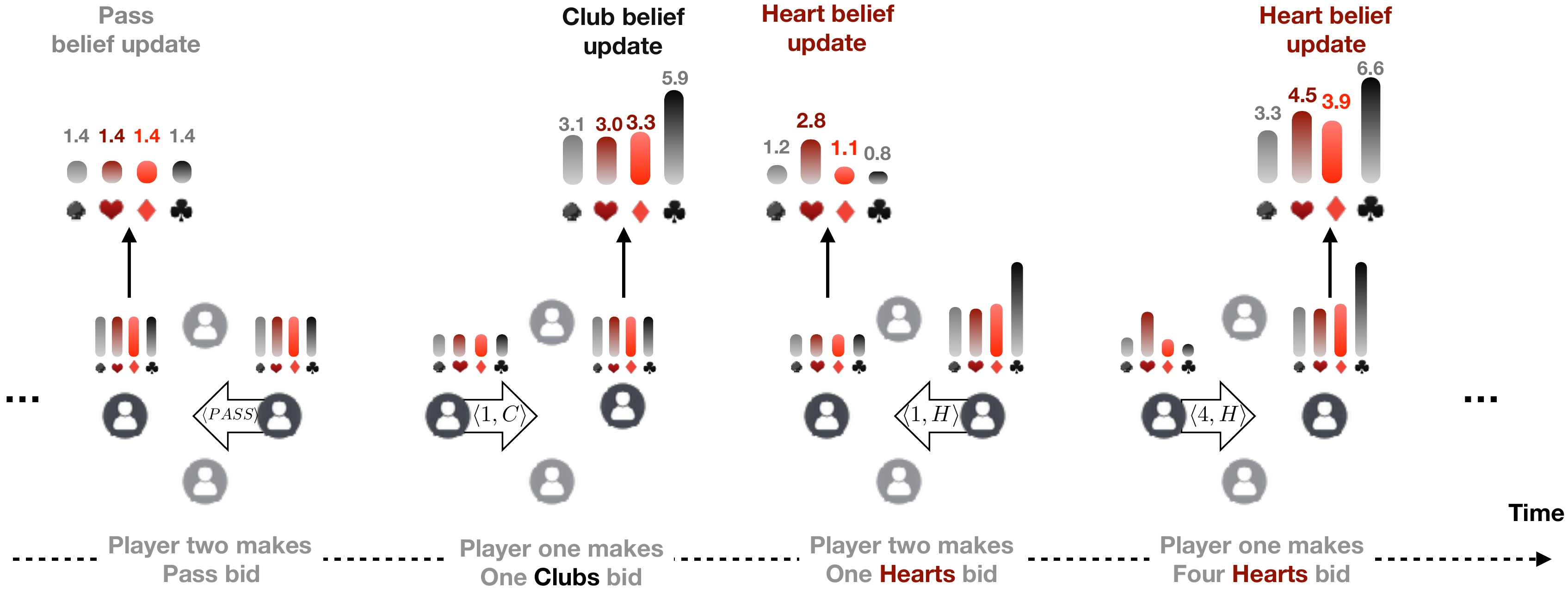}
\caption{An example of a belief update trace showing how PBL agents use actions for effective communication. Upon observing $\left\langle PASS\right\rangle$ from East, West decreases its HCP belief in all suits. When West bids $\left\langle 1, \text{C}\right\rangle$, East improves belief in clubs. Next, East bids $\left\langle 1, \text{H}\right\rangle$. West recalculates its belief from last time step and increases its HCP belief in hearts.}\label{fig:beliefupdate}
\end{figure*}

Non-competitive contract bridge bidding is an imperfect-information game that requires information exchange between agents to agree high-quality contracts. Hence, such a game serves as an ideal test-bed for PBL. In bridge, two teams of two (North-South vs East-West) are situated in opposing positions and play a trick-taking game using a standard 52-card deck.
Following a deal, bidding and playing phases can be effectively separated. During the bidding phase, players sequentially bid for a contract until a final contract is reached. A \textit{PASS} bid retains previously proposed contracts and a contract is considered final if it is followed by three consecutive \textit{PASS} bids. A non-\textit{PASS} bid proposes a new contract of the form $\left\langle \texttt{integer}, \texttt{suit} \right\rangle$, where $\texttt{integer}$ takes integer values between one and seven, and $\texttt{suit}$ belongs to $\{\clubsuit,\diamondsuit, \heartsuit, \spadesuit, \text{NT} \}$. The number of tricks needed to achieve a contract are $6+\texttt{integer}$, and an $\text{NT}$ suit corresponds to bidding to win tricks without trumps. A contract-declaring team achieves points if it fulfills the contract, and if not, the points for the contract go to the opposing team. Bidding must be non-decreasing, meaning $\texttt{integer}$ is non-decreasing and must increase if the newly proposed trump suit precedes or equals the currently bid suit in the ordering $\clubsuit < \diamondsuit < \heartsuit < \spadesuit < NT $.

In this work, we focus on non-competitive bidding in bridge, where we consider North (N) and South (S) bidding in the game, while East (E) and West (W) always bid \textit{PASS}. Hence, the declaring team never changes. Thus, each deal can be viewed as an independent episode of the game. The private information of player $i \in \{N, S\}$, $x^{i}$, is its hand. $x^i$ is a 52-dimensional binary vector encoding player $i$'s 13 cards. An agent's observation at time step $t$ consists of its hand and the bidding history: $o^{i}_t= \{x^{i}_{t}, h^{i}_t \}$. In each episode, Players N and S are dealt hands $x^N, x^S$ respectively. Their hands, together, describe the full state of the environment $s=\{x^N, x^S \}$, which is \emph{not fully observed} by either of the two players. Since rolling out via self-play for every contract is computationally expensive, we resort to double dummy analysis (DDA) \cite{haglund2010search} for score estimation. Interested readers are referred to~\cite{haglund2010search} and the appendix for further details. In our work, we use standard Duplicate bridge scoring rules~\cite{duplicatebridgelaw} to score games and normalize scores by dividing them by the maximum abusolute score.

\vspace{0.15cm}\noindent\textbf{Benchmarking \& Ablation Studies:} PBL introduces several building blocks, each affecting performance in its own right. We conduct an ablation study to better understand the importance of these elements and compare against a state-of-the-art method in PQL~\cite{Yeh16}. We introduce the following baselines:
\begin{enumerate}
    \item\textbf{Independent Player (IP)}: A player bids independently without consideration of the existence of the other player. 
    \item \textbf{No communication reward (NCR)}: One important question to ask is how beneficial the additional communication auxiliary reward $r_c$ is in terms of learning a good bidding strategy. To answer this question, we implement a baseline using the same architecture and training schedule as PBL but setting the communication reward weighting to zero, $\alpha = 0.$ 
    \item \textbf{No PBL style iteration (NPBI)}: To demonstrate that multiple iterations between policy and belief training are beneficial, we compare our model to a baseline policy trained with the same number of weight updates as our model but no further PBL iterations after training a belief network $\Phi_0$ at PBL iteration $k=0$.
    \item \textbf{Penetrative Q-Learning (PQL)}: PQL as proposed by Yeh and Lin \shortcite{Yeh16} as the first bidding policy for non-competitive bridge bidding without human domain knowledge.
    \end{enumerate}
\begin{figure*}[ht]
\centering
\begin{subfigure}{0.5\textwidth}
\centering
\includegraphics[width=\textwidth]{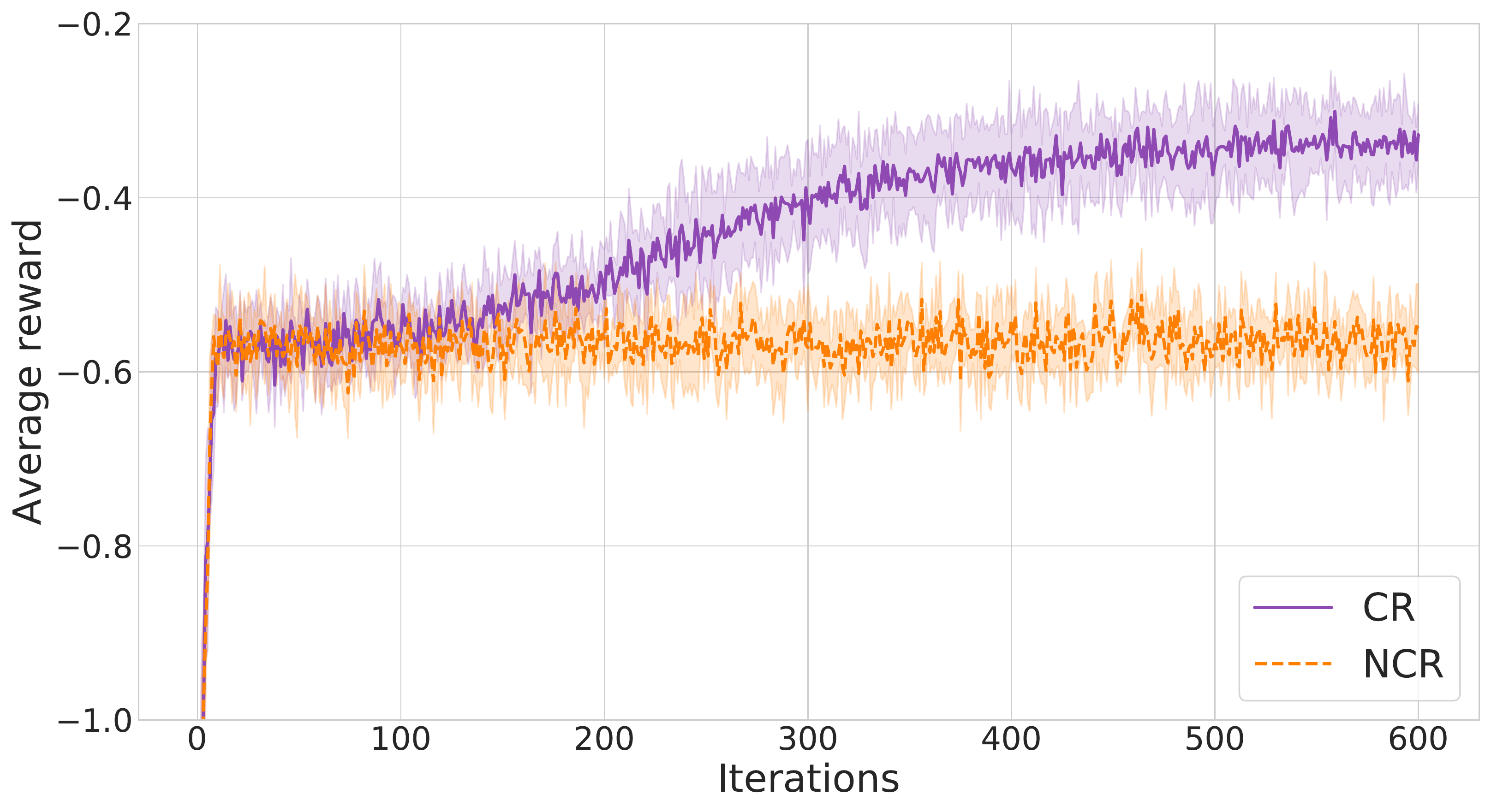}
\caption{Learning curves for Silent Guide}
\label{fig:guide_learning}
\end{subfigure}
~
\begin{subfigure}{0.23\textwidth}
\centering
\includegraphics[width=\textwidth]{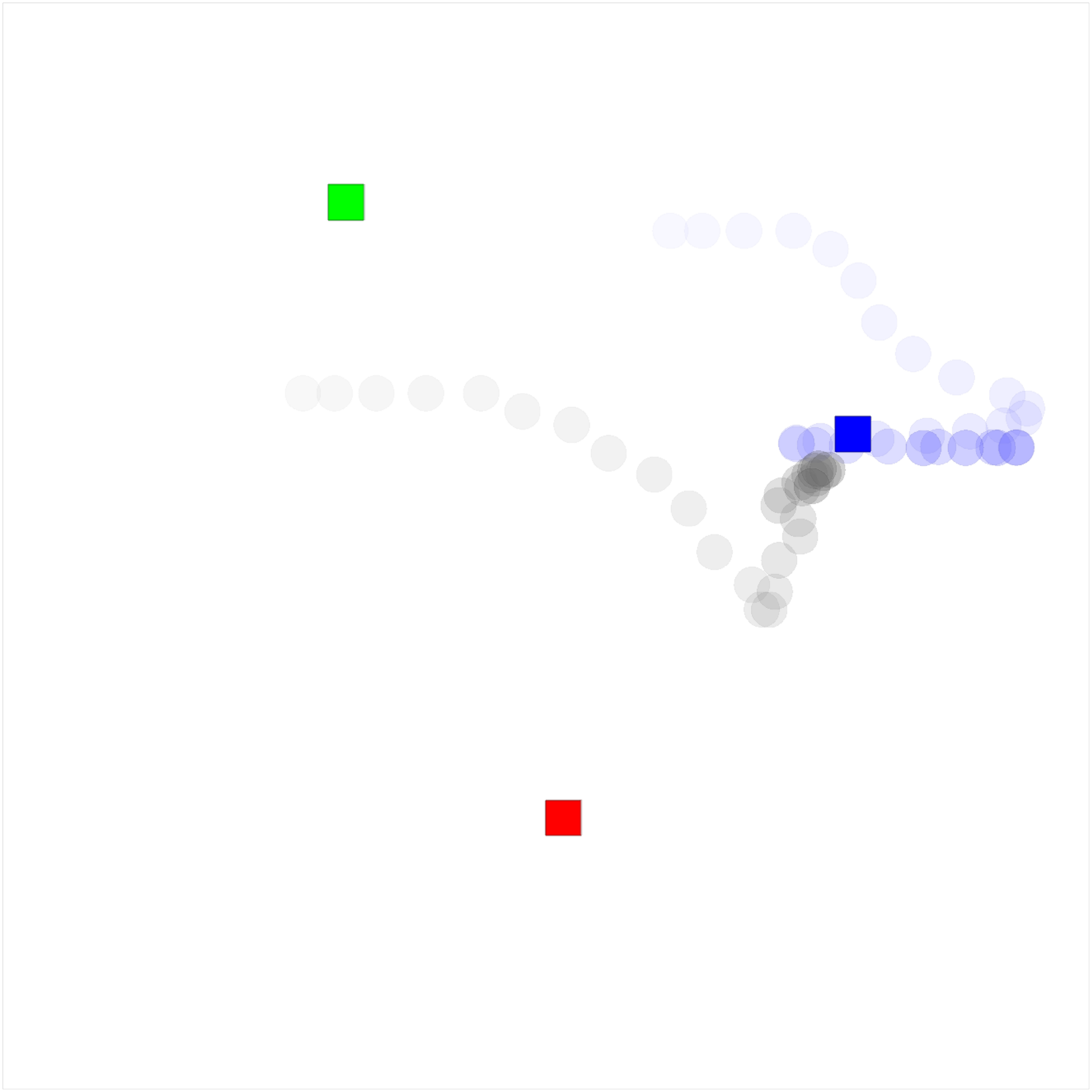}
\caption{CR}
\label{fig:guide_cr}
\end{subfigure}
~
\begin{subfigure}{0.23\textwidth}
\centering
\includegraphics[width=\textwidth]{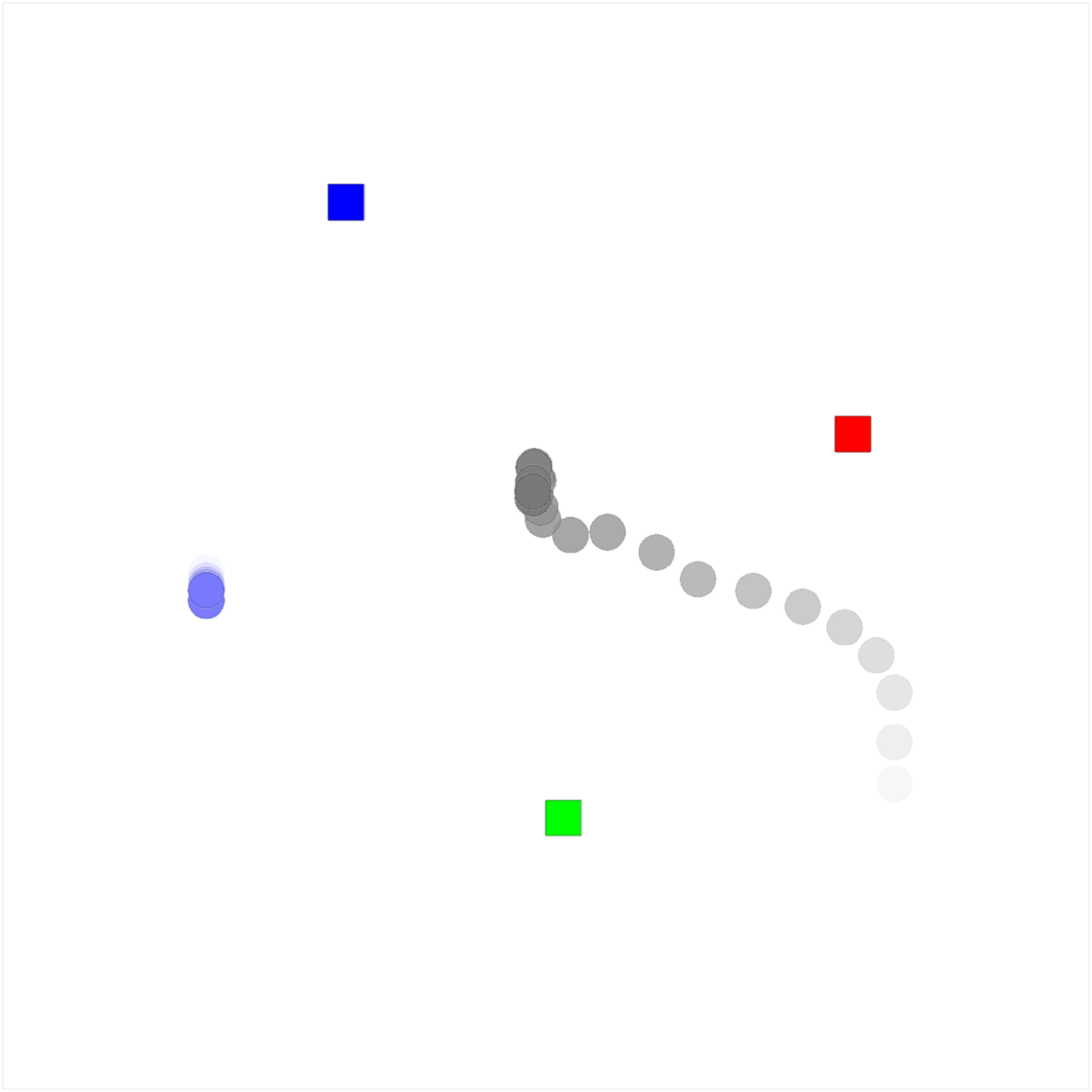}
\caption{NCR}
\label{fig:guide_ncr}
\end{subfigure}
\caption{a) Learning curves for Silent Guide. Guide agent trained with communication reward (CR) significantly outperforms the one trained with no communication reward (NCR). b) A trajectory of Listener (gray circle) and Guide (blue circle) with CR. Landmarks are positioned randomly and the Goal landmark (blue square) is randomly chosen at the start of each episode. c) A trajectory of Listener and Guide with NCR. Trajectories are presented with agents becoming progressively darker over time.}
\label{fig:silentguideresults}
\end{figure*}

Figure~\ref{fig:non-comp-bridge} shows the average learning curves of our model and three baselines for our ablation study. We obtain these curves by testing trained algorithms periodically on a pre-generated test data set which contains 30,000 games. Each point on the curve is an average score computed by Duplicate bridge scoring rules~\cite{duplicatebridgelaw} over 30,000 games and 6 training runs. As can been seen, IP and NCR both initially learn faster than our model. This is reasonable as PBL spends more time learning a communication protocol at first. However, IP converges to a local optimum very quickly and is surpassed by PBL after approximately 400 learning iterations. NCR learns a better bidding strategy than IP with a belief module. However, NCR learns more slowly than PBL in the later stage of training because it has no guidance on how to convey information to its partner. PBL outperforming NPBI demonstrates the importance of iterative training between policy and belief modules.

\noindent\textbf{Restrictions of PQL:} 
PQL~\cite{Yeh16} is the first algorithm trained to bid in bridge without human engineered features. However, its strong bidding performance relies on heavy adaption and heuristics for non-competitive Bridge bidding. First, PQL requires a predefined maximum number of allowed bids in each deal, while using different bidding networks at different times. Our results show that it will fail when we train a single NN for the whole game, which can been seen as a minimum requirement for most DRL algorithms. Second, PQL relies on a rule-based function for selecting the best contracts at test time. In fact, removing this second heuristic significantly reduces PQL's performance as reported in Figure~\ref{fig:pbl_vs_pql_bar}. In addition, without pre-processing the training data as in \cite{Yeh16}, we could not reproduce the original results. To achieve state-of-the-art performance, we could use these (or other) heuristics for our bidding algorithm. However, this deviates from the focus of our work which is to demonstrate that PBL is a general framework for learning to communicate by actions. 

\vspace{0.15cm}\noindent\textbf{Belief Update Visualization:} To understand how agents update their beliefs after observing a new bid, we visualize the belief update process (Figure \ref{fig:beliefupdate}). An agent's belief about its opponent's hand is represented as a 52-dimensional vector with real values which is not amenable to human interpretation. Therefore, we use \emph{high card points} (HCPs) to summarize each agent's belief. For each suit, each card is given a point score according to the mapping: \textbf{A=4, K=3, Q=2, J=1, else=0}. Note that while agents' beliefs are updated based on the entire history of its opponent's bids, the difference between that agent's belief from one round to the next is predominantly driven by the most recent bid of its opponent, as shown in Figure~\ref{fig:beliefupdate}.

\vspace{0.15cm}\noindent\textbf{Learned Bidding Convention:} Whilst our model's bidding decisions are based entirely on raw card data, we can use high card points as a simple way to observe and summarize the decisions which are being made. For example, we observe our policy opens the bid with $1\spadesuit$ if it has HCPs of spade $4.5$ or higher but lower HCPs of any other suits. We run the model on the unseen test set of 30,000 deals and summarize the learned bidding convention in Appendix \ref{appendix:hcpt}.

\vspace{0.15cm}\noindent\textbf{Imperfect Recall of History:} the length of the action history players can recall affects the accuracy of the belief models. The extent of the impact depends on the nature of the game. In bridge, the order of bidding encodes important information. We ran an ablation study where players can only recall the most recent bid. In this setting, players do worse (average score 0.065) than players with perfect recall. We conjecture that this is because players can extract less information and therefore the accuracy of belief models drops.

\subsection{Silent Guide}
We modify a multi-agent particle environment \cite{lowe2017multi} to test the effectiveness of our novel auxiliary reward in a distributed setting. This environment also allows us to explore the potential for implicit communication to arise through machine theory of mind. In the environment there are two agents and three landmarks. We name the agents Guide and Listener respectively. Guide can observe Listener's goal landmark which is distinguished by its color. Listener does not observe its goal. However, Listener is able to infer the meaning behind Guide's actions. The two agents receive the same reward which is the negative distance between Listener and its goal. Therefore, to maximize the cumulative reward, Guide needs to tell Listener the goal landmark color. However, as the ``Silent Guide" name suggests, Guide has no explicit communication channel and can only communicate to Listener through its actions.

In the distributed setting, we train separate belief modules for Guide and Listener respectively. The two belief modules are both trained to predict a naive agent's goal landmark color given its history within the current episode but using different data sets. We train both Guide and Listener policies from scratch. Listener's policy takes Listener's velocity, relative distance to three landmarks and the prediction of the belief module as input. It is trained to maximize the environment reward it receives. Guide's policy takes its velocity, relative distance to landmarks and Listener's goal as input. To encourage communication by actions, we train Guide policy with the auxiliary reward proposed in our work. We compare our method against a naive Guide policy which is trained without the communication reward. The results are shown in Figure~\ref{fig:silentguideresults}. Guide when trained with communication reward (CR) learns to inform Listener of its goal by approaching to the goal it observes. Listener learns to follow. However, in NCR setting, Listener learns to ignore Guide's uninformative actions and moves to the center of three landmarks. While Guide and Listener are equipped with belief models trained from different data sets, Guide manages to use its own belief model to establish the mental state of Listener and learns to communicate through actions judged by this constructed mental state of Listener. We also observe that a trained Guide agent can work with a naive RL listener (best reward -0.252) which has no belief model but can observe PBL guide agent’s action. The success of Guide with CR shows the potential for machine theory of mind. We obtain the learning curves by repeating the training process five times and take the shared average environment reward.

\section{Conclusions \& Future Work}
\label{future}
In this paper, we focus on implicit communication through actions. This draws a distinction of our work from previous works which either focus on explicit communication or unilateral communication. We propose an algorithm combining agent modeling and communication for collaborative imperfect-information games. Our PBL algorithm iterates between training a policy and a belief module. We propose a novel auxiliary reward for encouraging implicit communication between agents which effectively measures how much closer the opponent's belief about a player's private information becomes after observing the player's action. We empirically demonstrate that our methods can achieve near optimal performance in a matrix problem and scale to complex problems such as contract bridge bidding. We conduct an initial investigation of the further development of machine theory of mind. Specifically, we enable an agent to use its own belief model to attribute mental states to others and act accordingly. We test this framework and achieve some initial success in a multi-agent particle environment under distributed training. There are a lot of interesting avenues for future work such as exploration of the robustness of collaboration to differences in agents' belief models.

\bibliographystyle{aaai}
\bibliography{main}

\newpage
\onecolumn
\appendix
\setcounter{secnumdepth}{2}

\section{Bridge}
\label{appendix:bridge}
\subsection{Playing Phase in Bridge}
After the final contract is decided, the player from the declaring side who first bid the trump suit named in the final contract becomes Declarer. Declarer's partner is Dummy. The player to the left of the declarer becomes the first leading player. Then Dummy lays his cards face up on the table and then play proceeds clockwise. On each trick, the leading player shows one card from their hand and other players need to play the same suit as the leading player if possible; otherwise, they can play a card from another suit. Trump suit is superior to all other suits and, within a suit, a higher rank card is superior to lower rank one. A trick is won by the player who plays the card with the highest priority. The winner of the hand becomes the leading player for the next trick.

\subsection{Double Dummy Analysis (DDA)}
Double Dummy Analysis assumes that, for a particular deal, one player's hand is fully observed by other players and players always play cards to their best advantage. However, given a set $s=\{x^N, x^S \}$, the distribution of remaining cards for the two non-bidding players East and West is still unknown. To reduce the variance of the estimate, we repeatedly sample a deal $U$ times by allocating the remaining cards randomly to East and West and then estimate ${r}_e(s)$ by taking the average of their DDA scores,
\begin{equation}
    {r}_e(s) = \frac{1}{U}\sum^U_{u=1}{r}_e(x^N, x^S, x^E_u, x^W_u),
\end{equation} where $x^E_u, x^W_u$ are hands for East and West from the $u^\text{th}$ sampling respectively.
For a specific contract $a_t$, the corresponding score is given by $r_e(a_t|s) = {r}_e(x^N, x^S, a_t)$. In our work, we set $U=20$.

\subsection{Bridge Scoring}
\begin{algorithm}
\caption{Bridge Duplicate Scoring}\label{scorefunct}
\begin{algorithmic}
\STATE{\bfseries Score}{ ($tricks\_made, bid\_level, trump\_suit$)}
\STATE $T \gets trump\_suit$
\STATE $\delta \gets tricks\_made - (bid\_level+6)$
\STATE $score \gets 0$
\IF {$\delta\geq 0$} 
\STATE $score \gets score + bid\_level * scale_T + bias_T $ \COMMENT {Contract tricks}
\IF {score $\geq100$}
\STATE $score \gets 300$ \COMMENT {Game Bonus}
\ELSE
\STATE $score \gets 50$ \COMMENT {PARTSCORE}
\ENDIF
\IF{$\delta = 6$}
\STATE $score \gets score + 500$ \COMMENT {Slam bonus}
\ELSIF{$\delta = 7$}
\STATE $score \gets score + 1000$ \COMMENT {Grand Slam bonus}
\ENDIF
\IF{$\delta > 0$}
\STATE $score \gets score + \delta*scale_T$ \COMMENT {Over-tricks}
\ENDIF
\ELSE 
\STATE $score \gets score - bid\_level * 50 $ \COMMENT {Under-tricks}
\ENDIF

\end{algorithmic}
\end{algorithm}
Algorithm \ref{scorefunct} shows how we score a game under Duplicate Bridge Scoring rules. We obtain the average of $tricks\_made$ using Double Dummy Analysis \cite{haglund2010search} given the hands of players North and South $\{ x^N, x^S\}$, the declarer and the trump suit. The score function above has a scale and bias for each trump suit. The scale is 20 for $\clubsuit$ and $\diamondsuit$ and 30 for all others. Bias is zero for all trumps, except NT which has a bias of 10.

Note that \textit{Double} is only a valid bid in response to a contract proposed by one's opponents. Also, a \textit{Redouble} bid must be preceded by a \textit{Double} bid. In the non-competitive game, opponents do not propose contracts, so these options are naturally not included.

\subsection{\textit{Double Pass} Analysis} 
At the beginning of bidding, when two players both bid \textit{Pass} (\textit{Double Pass}), all players' hands are re-dealt and a new episode starts. If we ignore these episodes in training, a naive strategy emerges where a player always bids \textit{Pass} unless it is highly confident about its hand and therefore bids at level 1 whose risk is at the minimum. In this work, we are interested in solving problems where private information needs to be inferred from observed actions for better performance in a game. Therefore, this strategy is less meaningful and the opportunity cost of bidding \textit{Pass} could be high when a player could have won the game with high reward. To reflect this opportunity cost in training, we set the reward for Double \textit{Pass} as the negative of the maximum attainable reward given the players' hands: $r_{dp}({s}) = -\max({r}_e({s}))$. Therefore, a player will be penalized heavily by bidding \textit{Pass} if it could have obtained a high reward otherwise and awarded slightly if they could never win in the current episode. It is worthy of note that an initial \textit{Pass} bid can convey information; however, if it is followed by a second \textit{Pass} bid the game ends and hence no further information is imparted from a second \textit{Pass}.

We note that, by Duplicate Bridge Laws 77 and 22 of \cite{duplicatebridgelaw}, if the game opens with four \textit{Pass} bids all players score zero. In our setting, however, East and West play pass regardless of their hands and this will give North and South Player extra advantages.  Therefore, we use $r_{dp}$ to avoid results where North and South only bid where they have particularly strong hands; we discourage full risk averse behavior.

\section{Experiment Details in Bridge}
\label{appendix:exp}
\subsection{Offline Environment}
Generating a new pair of hands for North and South and pre-calculating scores for every possible contract at the start of a new episode during policy training is time inefficient. Instead, we pre-generate 1.5 million hands and score them in advance. Then we sample new episodes from this data set when we train a policy. We also generate a separate test data set containing 30,000 hands for testing our models.

\subsection{Model Architecture}
\begin{figure}
    \centering
    \includegraphics[scale=0.2]{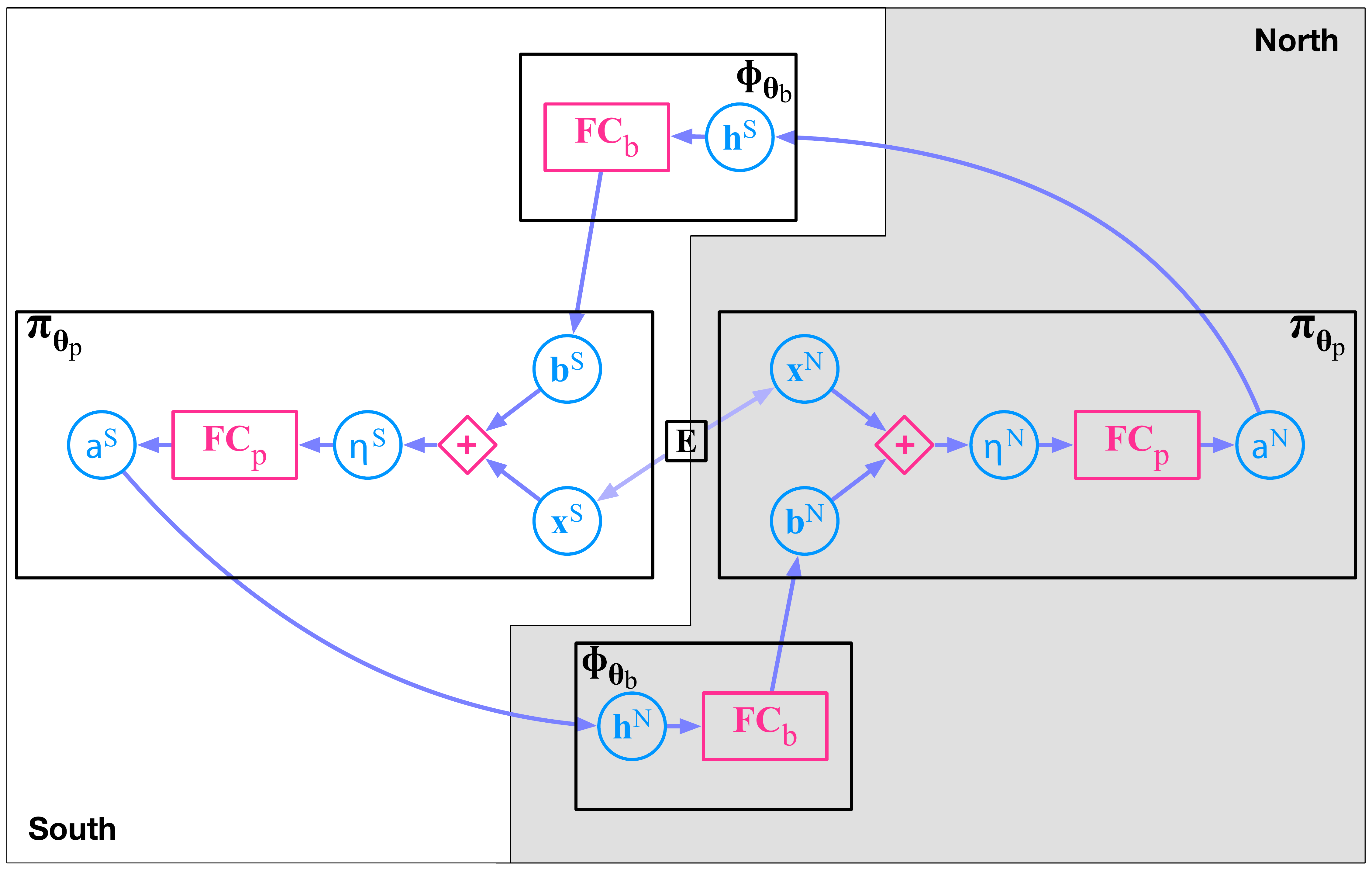}
    \caption{The architecture of our Policy Network and Belief Network in bridge. $\mathbf{E}$ represents environment and FC stands for fully connected layers. For player $i$, we add their private information $x^i$ and belief $b^s$ together to form $\eta^i$ as the input to its policy network $\pi$. Player $i's$ action $a^i$ becomes part of player $j$'s history $h^j$, which is the input to Player $j$'s belief network $\Phi$.}
    \label{fig:mdoel}
\end{figure}
We parameterize policy and belief modules with two neural networks $\pi_{\theta_p}$ and $\Phi_{\theta_b}$ respectively. Fig. \ref{fig:mdoel} shows our model for the bridge experiments. Weights of belief and policy modules are shared between players. The input to a player's policy $\pi_{\theta_p}$ is the sum of its private information ${x}$ and belief ${b}$. The last layer of the belief module $\Phi$ has a $\textit{Sigmoid}$ non-linear activation function, so that the belief vector ${b}$ has elements in the interval $[0, 1]$. By adding belief ${b}$ to ${x}$, we could reuse parameters associated with ${x}$ for ${b}$ and avoid training of extra parameters. 

\subsection{Policy Pre-training}\label{appendix:warmstart}
In the first PBL iteration $(k=0)$, to have a good initial policy and avoid one with low-entropy, we train our policy to predict a distribution formed using \text{Softmax} on ${r}^e({s})$ with temperature $\tau$. The loss for pre-train policy given ${s}$ is: $$\mathcal{L}_{k0}= \text{KL}(\pi({a}|{x})||\text{Softmax}({r}^e({s}), \tau)),$$ where $\text{KL}$ is the KL-Divergence. To have a fair comparison with other benchmarks, all our benchmarks are initialized with this pre-trained policy. Supervising a bidding policy to predict pre-calculated scores for all actions only provides it with a basic understanding of its hand.

\subsection{Policy Training}
We utilize Proximal Policy Optimization (PPO) \cite{schulman2017proximal} for policy training. Optimization is performed with the Adam optimizer \cite{kingma2014adam} with $\beta_1=0.9$, $\beta_2=0.999$, $\epsilon=10^{-8}$. The initial learning rate is $10^{-4}$ which we decay exponentially with a rate of decay of $0.95$ and a decaying step of $50$. 

The distance measure used in communication rewards $r^c$ is cross entropy and we treat each dimension of belief and ground truth as an independent Bernoulli distribution. 

We train PBL with 8 PBL iterations, we do policy gradient 200 times per iteration and we sample 5000 episodes each time. Each mini-batch is then a sub-sample of 2048 episodes. The initial communication weight $\alpha$ is $5$, and we decay it gradually. We train all other baselines with the same PPO hyperparameters.

\subsection{Learning A Belief Network}
When a player tries to model its partner's hand based on the observed bidding history, we assume it can omit the restriction that its partner can only hold 13 cards. Therefore, we take the prediction of the partner's hand given the observed bidding history as a 52-label classification problem, where each label represents one corresponding card being in the partner's hand. In other words, we treat each card from a 52-card deck being in the partner's hand as an independent Bernoulli distribution and we train a belief network by maximizing the joint likelihood of these 52 Bernoulli distributions given a bidding history $h$. This gives the loss for belief network as:
\begin{align}
    \mathcal{L}_{\Phi}= -\sum^{52}_{i=1} x^{-i}\log(b^i)+(1 - x^{-i})\log(1 - b^i)\nonumber,
\end{align}
where $x^{-i}$ and $b^i$ are elements of one-hot encoding vectors of a partner's hand $x^{-i}$ and one agent's belief $b^i$. The reasoning behind this assumption is we think it is more important to have a more accurate prediction over an invalid distribution than a less accurate one over a valid distribution as the belief itself is already an approximation.

For each iteration of belief training, we generate 300,000 data episodes with the current policy to train the belief network. Optimization is performed with the Adam optimizer with $\beta_1=0.9$, $\beta_2=0.999$, $\epsilon=10^{-8}$ and a decay rate of $0.95$. The initial learning rate is $10^{-3}$. The batch size is 1024. We split the data set such that $90\%$ of it is used for training and $10\%$ is for early stopping check to prevent overfitting.

\clearpage
\section{Learned Opponent Model and Bidding Convention}
\subsection{High Card Points Tables}

\label{appendix:hcpt}
\begin{table}[htb]
    \centering
    
    \begin{minipage}[t]{0.49\textwidth}
    \centering
    \begin{tabular}{cccccc}
    \toprule
    Bid &  $\clubsuit$ &  $\diamondsuit$ &  $\heartsuit$ &  $\spadesuit$ &  Total \\
    \midrule
    PASS            &          1.4 &             1.3 &           1.4 &           1.4 &    5.4 \\
    1$\spadesuit$   &          2.3 &             2.1 &           2.3 &  \textbf{4.5} &   11.2 \\
    1$\heartsuit$   &          2.3 &             2.2 &  \textbf{4.7} &           1.8 &   11.0 \\
    1$\diamondsuit$ &          2.2 &    \textbf{4.4} &           2.1 &           2.0 &   10.7 \\
    1$\clubsuit$    & \textbf{4.8} &             1.9 &           2.1 &           2.1 &   10.9 \\
    3NT             &          4.4 &             4.4 &           4.9 &           4.6 &   18.3 \\
    4$\diamondsuit$ &          3.0 &    \textbf{6.5} &           3.1 &           3.4 &   16.1 \\
    4$\clubsuit$    & \textbf{6.6} &             5.3 &           2.3 &           2.3 &   16.5 \\
    \bottomrule
    \end{tabular}
    \vspace{0.05cm}
    \caption{Opening bid - \textbf{own} aHCPs.}
    \label{tbl:openingBid}
    \end{minipage}
    ~
    \begin{minipage}[t]{0.49\textwidth}
    \centering
    \begin{tabular}{cccccc}
    \toprule
    Bid &  $\clubsuit$ &  $\diamondsuit$ &  $\heartsuit$ &  $\spadesuit$ &  Total \\
    \midrule
    PASS            &           1.4 &             1.3 &           1.3 &           1.3 &    5.3 \\
    1$\spadesuit$   &           2.3 &             2.1 &           2.2 &  \textbf{4.6} &   11.1 \\
    1$\heartsuit$   &           2.2 &             2.2 &  \textbf{4.7} &           1.9 &   11.0 \\
    1$\diamondsuit$ &           2.3 &    \textbf{4.5} &           2.0 &           2.0 &   10.8 \\
    1$\clubsuit$    &  \textbf{4.8} &             2.0 &           2.2 &           2.2 &   11.1 \\
    3NT             &           4.4 &             4.6 &           4.8 &           4.7 &   18.5 \\
    4$\diamondsuit$ &           2.8 &    \textbf{6.8} &           3.3 &           3.1 &   16.0 \\
    4$\clubsuit$    &  \textbf{6.2} &             4.0 &           3.0 &           2.8 &   16.0 \\
    \bottomrule
    \end{tabular}
    \vspace{0.05cm}
    \caption{\textbf{Belief} HCPs after observing opening bid.}
    \label{tbl:openingBelief}
    \end{minipage}
    \begin{minipage}[t]{\textwidth}
    \centering
    \begin{tabular}{cccccc}
    \\
    \toprule
    Bid &  $\clubsuit$ &  $\diamondsuit$ &  $\heartsuit$ &  $\spadesuit$ &  Total \\
    \midrule
    PASS            &           4.4 &             4.5 &           4.4 &           4.5 &   17.8 \\
    1$\spadesuit$   &           4.4 &             4.3 &           4.5 &  \textbf{7.1} &   20.3 \\
    1$\heartsuit$   &           4.6 &             4.4 &  \textbf{7.0} &           4.9 &   20.9 \\
    1$\diamondsuit$ &           4.3 &    \textbf{6.7} &           4.8 &           4.9 &   20.7 \\
    1$\clubsuit$    &  \textbf{7.2} &             4.3 &           4.7 &           4.9 &   21.1 \\
    2$\spadesuit$   &           4.8 &             4.7 &           5.2 &  \textbf{8.5} &   23.2 \\
    2$\heartsuit$   &           4.9 &             5.1 &  \textbf{8.6} &           5.1 &   23.7 \\
    3NT             &           6.6 &             6.6 &           7.1 &           7.2 &   27.6 \\
    4$\diamondsuit$ &           5.2 &    \textbf{8.5} &           5.5 &           5.6 &   24.9 \\
    4$\clubsuit$    &  \textbf{8.4} &             5.6 &           5.4 &           5.4 &   24.8 \\
    6$\heartsuit$   &           5.7 &             6.2 &  \textbf{9.9} &           6.9 &   28.7 \\
    6$\diamondsuit$ &           6.4 &    \textbf{9.2} &           7.3 &           7.1 &   30.0 \\
    6$\clubsuit$    & \textbf{10.8} &             4.0 &          10.0 &           6.5 &   31.3 \\
    6NT             &           7.5 &             7.0 &           8.7 &           9.0 &   32.2 \\
    \bottomrule
    \end{tabular}
    \vspace{0.05cm}
    \caption{Responding bid - \textbf{own + belief} aHCPs.}
    \label{tbl:respondingBid}
    \end{minipage}
\end{table}

Table \ref{tbl:openingBid} shows the average HCPs (aHCPs) present in a hand for each of the opening bidding decisions made by North. Once an opening bid is observed, South updates their belief; Table \ref{tbl:openingBelief} shows the effect which each opening bid has on South's belief. We show in  Table \ref{tbl:respondingBid} the responding bidding decisions made by South; aHCP values in table \ref{tbl:respondingBid} are the sum of HCPs in South's hand and South's belief over HCPs in North's hand. The values highlighted in bold for each row are the maximum values for the respective row. This is only done for rows where the bid is has a specified trump suit. 



\section*{Supplementary material}

## C.3   Interesting Bidding Examples

We note that in general, in the initial stages of bidding the suit of a bid is used to show strength in that suit with no trump bids used to show strength across all suits and *PASS* bids used to show a weak hand. In later stages of bidding, players either reiterate their strength with larger trick numbers in their bids or show strength in other suits. In the final rounds of bidding agents show their overall combined view of the pairing's hands to propose bids that reflect all 26 of the pair's cards. In the following examples, we describe how this bidding process manifests in certain cases. Note that in each case where a bid specifying the trump suit is made, the bidding pair have the majority of the cards in the trump suit in their collective hands. This is often essential for success in attaining suited bids.

### Example 1
**NORTH**

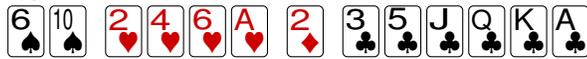

**SOUTH**

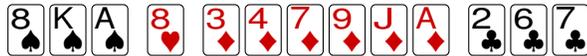

**BIDS**: 1♣ 1♢ 2♣ 2♢ 3NT PASS

**REWARD**: 0.283

North has a very strong hand in clubs holding all of the picture cards and the ace. North also holds the ace of hearts. South has strength in diamonds and spaces. The bidding pattern shows that North initially signals their strength in clubs and South their strength in diamonds. Both parties then reiterate their strengths in their next turns and a contract of 3NT is agreed.

### Example 2
**NORTH**

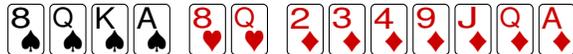

**SOUTH**

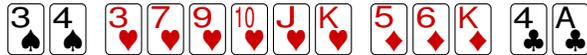

**BIDS**: 1♢ 1♡ 3NT 4♡ 6♡ PASS

**REWARD**: 0.301

North opens bidding with diamonds to show their strength in the suit. South then responds with hearts to show their strongest suit. North then bids no trumps to signal an inferred collective strength in multiple suits. South has little strength outside of hearts and continues to bid hearts and ultimately a contract of 6♡ is reached as bid by North who only has two hearts. This is a bid to win all but one hand with hearts as trumps which reflects the fact that between them the pair have from 7 to King of hearts. The pairing bid hearts despite their collective strength also in diamonds since hearts commands a higher per trick score if the contract is met.

### Example 3
**NORTH**

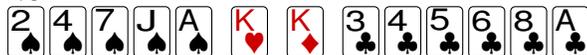

**SOUTH**

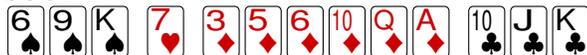

**BIDS**: 1♣ 1♢ 1♠ 2♣ 3NT 4♠ PASS

**REWARD**: 0.305

Between them the pairing have a strong hand in diamonds clubs and spades with a lack of hearts (15 + 13 = 28 HCP Total). Bidding opens with 1♣ as North aims to show their strength in clubs. South then conveys their strength in diamonds with a bid of 1♢. Bidding continues as North demonstrates some strength in spades with a bid of 1♠. Later, a bid by North of 3NT demonstrates that North believes the pair have strength across the board as reflected by the bids from all suits except hearts. Bidding ends however, with a bid of 4♠ from South which is then accepted through a *PASS* bid from North. South makes the bid of 4♠ despite only having 3 spades in their hand. The earlier bid of 1♠ from North informs South that North has some strength in spades and hence South believes that the pair may be able to attain 4♠.

## Example 4
**NORTH**
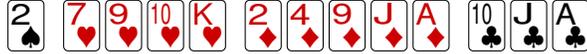
**SOUTH**
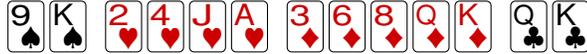

**BIDS**: 1♢ 3NT 4♡ 6♢ 6NT PASS

**REWARD**: 0.651

North has strength in all suits excluding spades (13 HCP Total) and South has strength evenly spread across all suits (18 HCP). Their shared strongest suit is diamonds where they hold the highest 4 cards between them. The bid of 6♢ is raised to 6NT as North forms a belief that the pair share strength across multiple suits (31 HCP Total for both hands).

## Example 5
**NORTH**
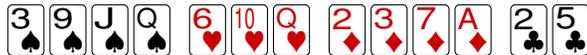
**SOUTH**
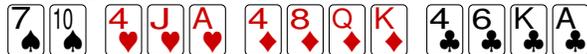

**BIDS**: PASS 1♣ 1♢ 2♢ 3NT PASS

**REWARD**: 0.211

North opens bidding with *PASS* signaling they they have a relatively weak hand (8 HCP Total). South then responds with 1♣ signaling their strength in clubs. Bidding then progresses as North bids 1♢ signaling that diamond is the stronger of their suits. South also has strength in diamonds and ultimately the pairing is able to reason about their collective hands and bid 3NT; a bid that would not be reached without forming beliefs due to the relative weakness of North's hand.

## Example 6
**NORTH**
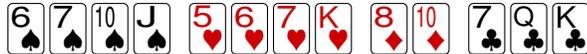
**SOUTH**
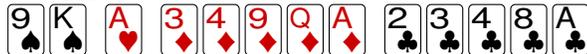

**BIDS**: PASS 1♣ 2♣ 2♢ 3NT PASS

**REWARD**: 0.258

Between them North and South have evenly spread strength across the suits (9 + 17 = 26 HCP Total). Similarly to Example 4, we see that the agents are able to reason, using their beliefs, about the collective strength of both of their hands. Here the pair reach a bid of 3NT while in Example 4, the pairing shared 31 HCP and reached a bid of 6NT.

## Example 7
**NORTH**
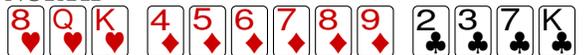
**SOUTH**
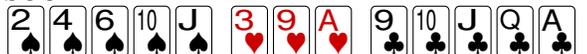

**BIDS**: PASS 1♣ 1♢ 1♠ 2♣ PASS

**REWARD**: 0.104

Both players lack any cards in one of the four suits. North opens bidding with PASS to show the relative weakness of their hand (8 HCP Total). Bidding remains low in different suits reflecting the similar level of strength across the suits while low bids show no significant strength overall. Ultimately a bid of 2♣ is reached reflecting strength in ♣ as between them the pair hold from 9 to Ace of clubs.



\end{document}